\documentclass{article}
\usepackage{graphicx} 
\usepackage{color}
\usepackage{amsmath}
\usepackage{mathtools}
\allowdisplaybreaks
\usepackage{amssymb}
\usepackage{bm}
\usepackage{physics}
\usepackage[mathscr]{euscript}
\usepackage{multirow}
\usepackage{enumitem}

\usepackage{xcolor}
\usepackage{tcolorbox}
\usepackage{tikz}
\usepackage{CJK}

\usepackage[colorlinks,citecolor=blue,linkcolor=blue,urlcolor=blue,pagebackref]{hyperref}
\usepackage{amsthm}
    \theoremstyle{plain} 

    \theoremstyle{definition}

    \theoremstyle{remark}
        
\usepackage[margin=1in]{geometry}
\usepackage{booktabs}
\usepackage{microtype}
\usepackage{setspace}
\usepackage[ruled]{algorithm2e}
\usepackage[labelfont=bf, font=small]{caption}
\usepackage{natbib}
    \bibpunct[, ]{(}{)}{,}{a}{}{,}%

\title{OpenAI-o1 AB Testing: Does the o1 model really do good reasoning in math problem solving?}
\author{Leo Li\thanks{Email: leol109828@gmail.com}, Ye Luo\thanks{HKU Business School, The University of Hong Kong. Email: kurtluo@hku.hk}, Tingyou Pan\thanks{The University of Hong Kong. Email: u3597480@connect.hku.hk}}
\date{October 2024}

\begin{document}

\maketitle

\begin{abstract}
    The Orion-1 model by OpenAI is claimed to have more robust logical reasoning capabilities than previous large language models. However, some suggest the excellence might be partially due to the model ``memorizing" solutions, resulting in less satisfactory performance when prompted with problems not in the training data. We conduct a comparison experiment using two datasets: one consisting of International Mathematics Olympiad (IMO) problems, which is easily accessible; the other one consisting of Chinese National Team Training camp (CNT) problems, which have similar difficulty but not as publically accessible. We label the response for each problem and compare the performance between the two datasets. We conclude that there is no significant evidence to show that the model relies on memorizing problems and solutions. Also, we perform case studies to analyze some features of the model's response.
\end{abstract}

\section{Background}

The OpenAI Orion-1 model, commonly referred to as o1, was unveiled on September 12th, 2024, and has garnered significant attention since its release. This model, particularly the o1-preview and o1-mini variants, has been lauded for its advanced reasoning and logical capabilities, particularly in the realms of mathematical problem-solving and coding.

One notable feature of the o1 models is their exceptional reasoning prowess, setting them apart from other Language Learning Models (LLMs) such as GPT-4o and Claude 3.5. The o1 models' superior reasoning abilities are attributed to a novel training approach that involves employing a token-wise reward model through reinforcement learning. By leveraging this methodology, the o1 models emulate the reasoning and reflective processes, thereby fostering an intrinsic chain-of-thought style in token generation. In essence, the o1 model endeavors to formulate plans and subsequently execute them in the process of solving mathematical problems and engaging in other cognitive tasks.

While the o1 models have demonstrated a strong level of reasoning capability, as highlighted by OpenAI's claim that o1-mini's score in the high school AIME math competition is around the top 500 US students, the evaluation of these models against a selection of private high school math questions yielded results that were not as impressive as anticipated. This discrepancy has prompted a research inquiry aimed at scientifically investigating whether the o1 models, akin to other Language Learning Models (LLMs), rely predominantly on memorizing solution steps in a chain of thought (CoT) fashion, or if they can genuinely generate high-quality reasoning steps even for unfamiliar math problems. The primary focus of this research is to assess the generalizability of the o1 models' problem-solving abilities. Specifically, the study aims to determine if the o1 models possess robust generalizability compared to LLMs like GPT-4o and whether their reasoning capabilities contribute to enhanced problem-solving generalization. Furthermore, the investigation seeks to identify the difficulty level at which the o1 models' generalization capabilities significantly surpass those of standard LLMs such as GPT-4o.

To carry out rigorous research steps towards answering the questions in the above, we adopt a A/B-test approach. On the one hand, we construct benchmark problem sets such as the problems in the International Mathematics Olympiad (IMO), which is considered as the highest level of math competition amongst high school students in mathematics, and these problems have good open access to all the LLMs including o1. On the other hand, we construct a comparable problem set that are considered to have similar difficulty level in mathematics, for example, the Chinese Mathematics Olympiad (CMO), that are less accessible to the LLMs during training. Of course, one can still argue that the CMO problems also have open access compared to IMO, but CMO problems are less accessible and less likely to be pre-trained by these LLMs. In the end, we construct a private set of problems in the level of high school mathematics, which is even less likely to be accessed by OpenAI, with similar difficulty level compared to IMO problems. These problems are collected from the Chinese National Team Training camp (CNT), which aims at selecting representative students to participate IMO each year in China. Comparative statistics between IMO dataset and CNT datasets are provided to investigate on the performance of the o1 model.

Another approach we adopt after the large scale statistical investigation and tests, is to perform case study on a few presentative problems. We provide both positive ones and negative ones, with different types of pros and cons from a mathematical/logical reasoning perspective.

The case studies show that the o1 model has a prevalent problem with the ``search" type of problems. The definition of such a problem and an example is in Section 2. In most situations, the model can only search for possible valid cases with small numbers or simple functions. Also, for all kinds of questions, the model sometimes provides skeletal problem-solving steps similar to a correct solution a human contestant would write, but they lack the crucial detailed justification in between. In other situations, the model may provide incorrect intuition.

This research contributes to two primary areas. First, it advances the evaluation of AI systems' capabilities in mathematical reasoning and automated theorem proving. Second, it offers an in-depth assessment of OpenAI's o1 LLM performance across a broad spectrum of complex reasoning tasks, specifically in the categories of ``search," ``solve," and ``prove" for various types of mathematical problems. These contributions provide a comprehensive understanding of the model's abilities in addressing different aspects of mathematical reasoning.

A substantial body of research has focused on evaluating the mathematical reasoning capabilities and automated theorem proving performance of AI systems. Large Language Models (LLMs) have demonstrated strong proficiency across a wide range of mathematical benchmarks, spanning from elementary-level problems, as shown in \cite{cobbe2021training}, \cite{yu2023metamath}, and \cite{zhou2023solving}, to more advanced high-school and pre-university level challenges, as evidenced by \cite{mao2024champ}, \cite{urrutia2023s}, and \cite{hendrycks2021measuring}. Moreover, LLMs have been tested on highly complex tasks, such as International Mathematical Olympiad (IMO)-level problems. For instance, \cite{sinha2024wu} and \cite{trinh2024solving} specifically address the solution of IMO Geometry problems, while \cite{he2024olympiadbench} and \cite{huang2024olympicarena} introduce a range of Olympic-level problems, including mathematical problems, to evaluate AI systems' problem-solving capabilities. Notably, in the OlympicArena, the highest performance was achieved by GPT-4o, with an overall accuracy of $39.97\%$, whereas other models struggle to surpass the $20\%$ threshold. In the Olympiadbench, GPT-4's accuracy for solving IMO problems is $17.97\%$. Additionally, LLMs are being increasingly employed in the domain of automated theorem proving, as illustrated by systems such as LeanDojo \cite{yang2024leandojo} and LeanAgent \cite{kumarappan2024leanagent}, which leverage AI for formal theorem proving in mathematics.

The discussion surrounding what LLMs can do and how well they perform has been extensive. Numerous studies have systematically analyzed the performance of various types and versions of LLMs across different tasks. For instance, works like \cite{chang2024survey}, \cite{zhao2023survey}, and \cite{minaee2024large} provide comprehensive discussions on the properties, contributions, and limitations of popular LLM families (GPT, LLaMA, PaLM). Compared to previous versions of GPT, the o1 model demonstrates exceptional logical reasoning abilities and a broad knowledge base across multiple fields. \cite{wu2024comparative} explores the reasoning patterns of the o1 model in math, code, and commonsense reasoning, while \cite{gui2024logicgame} introduces a LogicGame designed to assess the reasoning abilities of various LLMs, including the o1 model. \footnote{Notably, the results from LogicGame show that even the o1 model's reasoning score remains below the $50\%$ threshold.} Additionally, \cite{xie2024preliminary} and \cite{hu2024can} discuss the potential applications of LLMs like o1 in fields such as medical assistance and code debugging. For a general evaluation of the o1 model, see \cite{zhong2024evaluation}.

Our research, however, focuses specifically on evaluating the performance of OpenAI's o1 model in the domains of mathematical reasoning and automated theorem proving. According to \cite{qiao2024we}, many LLMs rely heavily on Rote Memorization (RM), where pre-training data includes exact matches to problems found in the model’s memory, leading to a pattern where solving mathematical problems resembles memory retrieval rather than genuine reasoning. However, GPT-4 shows a stronger reliance on Inadequate Generalization (IG) over RM in mathematical reasoning. In our analysis, we demonstrate that the o1 model exhibits a reduced dependence on RM, indicating a shift towards more effective generalization.

By focusing on the o1 model's performance in these specific tasks, our study aims to deepen the understanding of LLMs' applicability and limitations in complex reasoning tasks, as well as to investigate how LLMs approach and solve mathematical problems.

\section{Setup of Our Tests}\label{sec:Setup}

To compare the difficulty levels of math problem sets with varying accessibility, two datasets were compiled for analysis. The first dataset consists of 60 problems sourced from the International Mathematical Olympiad (IMO) over the past decade. The IMO is a prestigious international competition held annually in July, where students from around the world participate in solving challenging mathematical problems. The second dataset comprises 60 problems from the Chinese National Team (CNT) training camp, which is not as readily available to the public as the IMO dataset.

The CNT training camp is a rigorous program designed to prepare students for the IMO competition in China. It takes place in the spring of each year and serves as a crucial training ground for aspiring mathematicians. The selection process for the Chinese national team involves multiple tests (typically 8-10 times), each lasting 4 and a half hours, mirroring the format of the actual IMO competition. These tests are considered mock exams for the IMO, aimed at evaluating the trainees' problem-solving skills, mathematical abilities, and consistency in performance under exam conditions.

The two data sets are consider to have privacy levels that:
\begin{equation*}
    \textrm{Privacy of IMO Data} < \textrm{Privacy of CNT Data}.
\end{equation*}

The null hypothesis posits that the o1-mini's problem-solving capability is based on reasoning skills. On the other hand, the alternative hypothesis suggests that the o1-mini's performance may be attributed to memorization of problems and solutions, or the imitation of pre-trained patterns. In the context of the null hypothesis, we would expect to observe comparable or similar performance levels in both the International Mathematical Olympiad (IMO) and Chinese National Team (CNT) datasets. Conversely, under the alternative hypothesis, we anticipate significant performance discrepancies between the two datasets, with the IMO dataset showing markedly higher performance levels. This difference is attributed to the likelihood of the o1-model having memorized the IMO dataset due to its inclusion in the pre-trained data of the Language Learning Model (LLM). Expanding upon this discussion, it is essential to consider the implications of these hypotheses on the evaluation of the o1-mini's problem-solving abilities. If the null hypothesis is supported, it would suggest that the o1-mini's performance is driven by its capacity for logical reasoning and problem-solving strategies. This would imply that the model is capable of generalizing its reasoning skills across different problem sets, regardless of their origin or complexity. Conversely, if the alternative hypothesis is upheld, it would raise questions about the o1-mini's reliance on memorization or pattern recognition rather than genuine problem-solving abilities. In this scenario, the model's performance in the IMO dataset, which is more likely to have been memorized due to its presence in the pre-trained data of the LLM, would be artificially inflated compared to its performance in the CNT dataset. This would underscore the importance of distinguishing between true problem-solving prowess and the ability to regurgitate memorized solutions. In conclusion, the investigation of these hypotheses not only sheds light on the o1-mini's problem-solving mechanisms but also underscores the need for robust evaluation methods to discern between reasoning-based performance and memorization-driven outcomes in artificial intelligence models.

As latex is a standard format of math problem writing and editing software, the three datasets are translated from PDF to latex files, so that the LLMs such as o1 can easily read and process. As many reports on o1 mention that no additional prompts such as CoT are needed, we directly feed these latex files of problems to the o1-mini model\footnote{while o1-mini is faster and cheaper than o1-preview model, reports \cite{openai} show that its mathematical capability is even slightly stronger than o1-preview, a much large LLM in parameter size.}. 

\textbf{Evaluation of Test Results}

To assess the test results, we utilize the standard grading methodology employed in mathematical competitions such as the International Mathematical Olympiad (IMO) or the Chinese National Team (CNT), where each problem is allocated a maximum of 7 points. When a problem requires a numerical answer, 1 point is awarded for providing the correct numerical solution. Additionally, if the intuitive approach to solving the problem is correct, 2 points are granted. The remaining 4 points are reserved for demonstrating meticulous and accurate reasoning steps, emphasizing the importance of detailed logical progression in mathematical problem-solving. In the realm of rigorous mathematics, the intricacies of reasoning and the precision of logical steps hold significant weight, while the overarching conceptual understanding, often favored by Language Learning Models (LLMs), receives comparatively less emphasis in the grading process.

For problems that are proof-oriented, the grading system involves awarding 2 points for a primarily correct chain of thought (CoT), indicating a logical pathway towards the solution. The remaining 5 points are contingent upon the presentation of detailed and rigorous arguments to substantiate the proof, underscoring the necessity of thorough and coherent reasoning in mathematical proofs.

During the evaluation of responses provided by the o1-mini, it was observed that the model struggles to deliver rigorous proof steps consistently. In contrast to formal proofs, the o1-mini often exhibits a trial-and-error approach, where a series of attempts are made, occasionally leading to correct answers through heuristic ``guesswork" facilitated by informal reasoning. This informal reasoning lacks the rigor and formality expected in mathematical proofs, highlighting a gap in the o1-mini's ability to consistently provide logically sound and detailed proof steps.

The presence of informal reasoning and heuristic ``guessing" in the o1-mini's problem-solving approach underscores the importance of enhancing the model's ability to engage in formal and rigorous mathematical reasoning. Addressing this limitation could lead to improved performance in providing structured and logical proofs, aligning more closely with the standards of mathematical competitions that prioritize meticulous reasoning and coherent argumentation in problem-solving processes.

An example of o1-mini ``guessing" the answer by verifying some cases with only small natural numbers involved is demonstrated below. The full response of o1-mini is included in the appendix.

\begin{tcolorbox}[colback = white, colframe = blue!50!white, title = Example of O1 ``Guessing" the Answer]

{\color{blue!50!white}\textbf{Problem:}}

Find all positive integers $a$, $b$, $c$ such that $ab-c$, $bc-a$, $ca-b$ is a power of $2$ (possible including $2^0 = 1$)

{\color{blue!50!white}\textbf{Brief Summary of o1-mini's solution:}}

O1 approached the problem by first fixing $a$ to be some integer, then letting
\[\begin{cases}
    ab - c = p \\
    ac - b = q
\end{cases}\]
where $p$, $q$ are some powers of $2$, to obtain a set of linear equations with two unknowns $b$ and $c$. After that, o1-mini solved the set of equations to get the value of $b$ and $c$ expressed in terms of $p$ and $q$. Finally, o1-mini plugged in various values to $p$ and $q$ to determine which tuples of $(a,b,c)$ are valid. O1-mini correctly identified all possible triples of $(a,b,c)$ in the end.

\end{tcolorbox}

Under the revised labeling standard, the evaluation process for the o1-mini's performance focuses on assessing its ability to demonstrate correct intuition and arrive at the correct results through reasoning, even in the absence of formal proofs. The new labeling criterion categorizes problems based on their nature into two distinct types:
\begin{enumerate}
    \item Search Type Problems: These problems require finding numerical, integer, or expression-based solutions of a specific kind. Below is an example of such questions. The o1-mini's success in correctly identifying such solutions through intuitive reasoning or ``guessing" is evaluated.

    \begin{tcolorbox}[colback = white, colframe = blue!50!white, title = Example of ``Search" Type Problem]
Find all functions $f: \mathbb{R} \Rightarrow \mathbb{R}$ that satisfies: For any real number $x, y$, the following recollectible sets are equal: $ \{f(x f(y)+1), f(y f(x)-1)\}=\{x f(f(y))-1, y f(f(x))+1\} $
    \end{tcolorbox} 

    \item   Solve Type Problems: These problems involve finding solutions to equations or optimization problems. The o1-mini's performance in deriving solutions to these types of problems is assessed based on its ability to reason and arrive at correct answers.
\end{enumerate}

The grading process is conducted meticulously by human evaluators proficient in the relevant mathematical domain. All problem sets, grades, and corresponding labels are available for review upon request, ensuring transparency and accessibility of the evaluation outcomes.

By focusing on evaluating the o1-mini's problem-solving capabilities in the context of ``Search" and ``Solve" type problems, the revised labeling standard aims to capture the model's proficiency in intuitively reasoning and generating correct solutions, even in the absence of formal proof steps. This approach provides a nuanced assessment of the o1-mini's performance, highlighting its capacity to navigate mathematical challenges through logical reasoning and problem-solving strategies.

\section{Analysis of Main Results}

Based on the information provided in the setup section, a dataset consisting of 60 problems from recent International Mathematical Olympiads (IMO) and 60 problems from the Chinese National Team training camp has been compiled for evaluation purposes. The evaluation methodology outlined in the setup section emphasizes assessing the o1-mini's ability to provide correct answers informally, focusing on the correctness of solutions rather than formally scrutinizing all reasoning steps. Upon examination of multiple cases, it has been observed that the o1-mini's problem-solving approach is characterized by a strong capacity for intuitive reasoning and the formulation of effective strategies to identify specific solutions, whether numerical or algebraic in nature. While the model may face challenges in delivering logically complete proofs, its strength lies in the ability to leverage intuition and strategic thinking to arrive at correct solutions within the given problem scenarios. This distinction underscores the o1-mini's proficiency in navigating mathematical challenges through intuitive reasoning and strategic problem-solving approaches, emphasizing its capability to excel in identifying specific solutions effectively, even in instances where formal proof construction may present challenges. By focusing on the model's strengths in intuition-driven solution finding, the evaluation process aims to provide insights into the o1-mini's problem-solving capabilities within the context of the IMO and Chinese National Team training camp problem sets.

\begin{table}[ht]
    \centering
    \begin{tabular}{|c|c|c|c|c|}
    \hline
       Problem Sets     & Total Quantity & Proof & Search & Solve \\ 
       \hline
        IMO & 60 &23 & 23& 14\\
        \hline
        CNT & 60 &10 & 27 & 23\\
        \hline
    \end{tabular}
    \caption{Distributions of problems }
    \label{tab:my_label}
\end{table}

Note that overall IMO and CNT problem sets have similar distributions of the problems of the three kinds, except that in CNT, the number proof problems are less than the number of proof kind of problems than IMO.

Our key evaluation metric is to examine whether o1-mini is able to offer the right answers in the Search and Solve types of problems. The results are summarized in the Table \ref{table:main-summary}.

\begin{table}[ht]
    \centering
    \begin{tabular}{|c|c|c|c|c|}
    \hline
    Problem Set & Search & Solve & Total & Benchmark (GPT-4o)\footnote{The GPT-4o result is tested in a dataset with lower level of difficulty. See {\color{red} add references} for reference.}\\
    \hline
       IMO  & 16 out of 23 (69.6\%) & 3 out of 14 (21.4\%) & 19 out of 37 (51.4\%) & 39.97\%\\
       \hline
       CNT  & 19 out of 27 (70.4\%) & 5 out of 23 (21.7\%) & 24 out of 50 (48\%) & 39.97\% \\
       \hline
       T-Stats in Diff & -0.0088  & -0.0038 & 0.0335 & \\
       \hline
    \end{tabular}
    \caption{Evaluation Results}
    \label{table:main-summary}
\end{table}

The first column demonstrates the empirical accuracy ratio
of the results that o1-mini gives in the 23 IMO problems of the kind of search, and than 27 IMO problems of the kind of search. Denote $p_{IMO}$ and $p_{CNT}$ are the accuracy ratio of o1 in the corresponding dataset. The t-statistics is computed by the following standard formula:
\begin{equation*}
    T-stats = \frac{p_{IMO} - p_{CNT}}{\sqrt{p_{IMO}(1-p_{IMO})} + \sqrt{p_{CNT}(1-p_{CNT})}}.
\end{equation*}
The t-statistics for both the ``Search" type and ``Solve" type problems are found to be insignificant and very close to 0. This outcome indicates that there is no statistically significant difference in the performance of the o1-mini model between the public dataset (IMO) and the private dataset (CNT). These results provide evidence to reject the hypothesis that the o1-mini model performs better on public datasets, suggesting that the model's capability is not derived from simply memorizing solutions but rather from its reasoning abilities.

Therefore, the findings support the argument that the o1-mini's proficiency in problem-solving stems from its \textbf{reasoning skills rather than from potential data leaks or reliance on memorized information.} The similarity in performance across public and private datasets indicates a consistent level of reasoning capability exhibited by the o1-mini model, reinforcing the notion that its problem-solving prowess is rooted in its ability to reason and strategize effectively rather than relying solely on pre-existing data or memorization.

\section{Case Study: Typical Cases of o1's Response}

This section investigates typical phenomena present in o1's responses. O1 usually outputs a thinking process written in narrative style and a final solution written in mathematically rigorous language. The intuition provided in the thinking process might be a crucial step in some situations. An example of o1 providing useful intuition is given here. Also, logical errors which accentuates itself in the final solutions part, such as failing to argue other solutions do not exist when answering a ``search" type question, are prevalent. The problem and a brief discussion o1-mini's performance is included in each case. O1-mini's full response is included in the first part of the appendix.

\subsection{Case 1: O1 providing correct intuition}

\begin{tcolorbox}[colback=white, colframe=blue!50!white, title=Problem 1]

A \emph{site} is any point $(x,y)$ in the plane for which $x,y \in \{1, \dots, 20\}$. Initially, each of the $400$ sites is unoccupied. Amy and Ben take turns placing stones on unoccupied sites, with Amy going first; Amy has the additional restriction that no two of her stones may be at a distance equal to $\sqrt5$. They stop once either player cannot move. Find the greatest $K$ such that Amy can ensure that she places at least $K$ stones.

The correct answer is 100.

\end{tcolorbox}

\textbf{Brief Summary and Comments on O1-mini's response:}

O1-mini correctly identified that only when two stones have a difference in coordinates of $(1,2)$ or $(2,1)$ can they have a distance of $\sqrt{5}$. Then, O1-mini noticed if one colors the points with black and white colors alternatively, as shown in the below picture, no two points of the same color may have a distance of $\sqrt{5}$. Finally, O1-mini concluded that if Amy adheres to putting her stones on the points with the same colors, she can place her stones on at least half of the same-colored points, whose number is $100$, regardless of her opponent's choices. In this example, o1-mini provided helpful intuition. Even though $100$ is the correct answer, O1-mini lacked an argument for why Amy couldn't guarantee more.

\begin{center}
    \begin{tikzpicture}
        \foreach \x in {0,1,...,4} {
            \foreach \y in {0,1,...,4} {
                \ifnum\intcalcMod{\x+\y}{2}=0
                    \fill[black] (\x*0.5, \y*0.5) circle (2pt);
                \else
                    \fill[white] (\x*0.5, \y*0.5) circle (2pt);
                    \draw (\x*0.5, \y*0.5) circle (2pt);
                \fi
            }
        }
    \end{tikzpicture}
\end{center}

\begin{center}
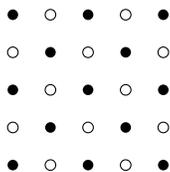
     Figure 1: Alternatingly coloring points so no two points of the same color have distance $\sqrt{5}$
\end{center}

\subsection{Case 2: O1-mini guessing the answer without arguing other answers do not exist}

\begin{tcolorbox}[colback=white, colframe=blue!50!white, title=Problem 2]

Determine all composite integers $n>1$ that satisfy the following property: if $d_1 < d_2 < \dots < d_k$ are all the positive divisors of $n$ with then $d_i$ divides $d_{i+1} + d_{i+2}$ for every $1 \leq i \leq k - 2$.

The correct answer is powers of primes.

\end{tcolorbox}

\textbf{Brief Summary and Comments on O1-mini's Response:}

O1-mini tested integers from $1$ to $18$, then selected several larger numbers. By analyzing the numbers that satisfy the given statement, it discovered the pattern that only powers of prime numbers are feasible. Then, o1-mini correctly proved why powers of prime numbers are feasible generally. However, when arguing why other composite numbers do not work, o1-mini only provided some examples to demonstrate this. In this problem, o1-mini adhered to testing small, easy-to-compute cases, an approach prevalent in most ``search" types of questions. Even though it is unable to explain why other untested cases are not feasible answers, which is usually part of the solution with the most points allocated, o1-mini remains capable of discovering patterns and then providing simple proofs.

\section{Case Study: Comparison of O1 Reasoning and Human Reasoning}

This section compares the logical reasoning process between human contestants and the o1 model. Correct methods of approaching and solving the question is presented in the Human Reasoning Process part of each subsection, followed with the Model's Reasoning Process section, which includes similarities to the human reasoning process and the o1 response's flaws. O1-mini's full response is included in the first part of the appendix.

\subsection{Case 3: Sum Divisibility Problem}
\begin{tcolorbox}[colback=white, colframe=blue!50!white, title=Problem 3]

Find all real numbers \( \alpha \) such that, for every positive integer \( n \), the integer 
\[
\left\lfloor \alpha \right\rfloor + \left\lfloor 2 \alpha \right\rfloor + \left\lfloor 3 \alpha \right\rfloor + \dots + \left\lfloor n \alpha \right\rfloor 
\]
is divisible by \( n \).

The correct answer is that \( \alpha \) must be even.

\end{tcolorbox}

\subsubsection{Human Reasoning Process}
One example of a ``human" reasoning process for solving this question involves first considering the case where $\alpha$ is an integer and evaluating two subcases, one where $\alpha$ is even and another when $\alpha$ is odd. It first equates the sum to $\frac {n(n+1)\alpha}{2}$ using the summation formula for arithmetic progressions. It is then inferred that for some odd $\alpha \frac {\alpha} {2}$ is not an integer, and therefore the expression is not divisible by n. 

Then the case where $\alpha$ is not an integer is considered. $\alpha$ is written as the sum of an integer $b$ and a fractional component $f$ where the integer is the floor of $\alpha$ and the fractional component is the difference between $\alpha$ and the integer such that $0<f<1$. $\alpha$ can be expressed as $b+f$, and the result is an equation which can then be manipulated. The floor of $k\alpha$ for some integer $k$ is equal to the sum of $kb$ and the floor of $kf$, and the original sum can be rewritten in terms of $b$ and $f$. 

Then 2 subcases are considered, one where $b$ is even and another where $\alpha$ is an odd integer plus a fractional component, in which $\alpha$ can be expressed as an even integer $b$ minus a fractional component $g$ which is the difference between 1 and the fractional component that is added to an odd number to get $\alpha$. It is then determined that the expression is not divisible by $n$ when $n=m$ and $m$ is the smallest integer larger than 1 such that the floor of $mf$ is larger than one for the first case, and $n=m$ and $m$ is the smallest integer larger than 1 such that the floor of $mg$ is larger than 1 for the second case. The expression, in these cases, will be 1 more and 1 less, respectively, than the sum of $n$ even integers which are divisible by $n$, meaning that the expressions will not be divisible by $n$. It can therefore be concluded that for non-integers $\alpha$ the expression will not be divisible by $n$ for all positive integers $n$.


\subsubsection{Model's Reasoning Process}
The model begins similarly, first considering the case where $\alpha$ is an integer $m$ and inferring that $m$ must be even for the expression to be true. 

However, the reasoning, as the model presents it, is flawed, as it first states, without specifying that $m$ is even, that the expression $\frac{m n(n+1)}{2}$ is clearly divisible by n (a false premise), and thus the expression is divisible by n if and only if $m$is even, which is not a valid inference.

It still reaches the correct conclusion from a separate, valid argument, asserting that $\frac {m(n+1)}{2}$ must be an integer for all $n$ for the expression to be divisible by $n$, which is only the case when $m$ is even, so the previous argument is likely the result of an error in expression of the reasoning, not the reasoning itself.

The model then evaluates the expression for non-integer $\alpha$, again similarly to the human solution, and it replicates perfectly the steps of the human solution up till where the sum is rewritten in terms of the floor of $\alpha$ and the fractional component. Instead of looking at individual subcases and determining the values for which the expressions aren’t divisible by $n$, the model rewrites the expression as the sum of the sum of the integer component and the sum of the fractional component.
It examines the sum of the fractional component, which is $\sum_{k=1}^n \lfloor k \delta \rfloor$ where $\delta$ is the fractional component, which it determines must be congruent to 0 modulo $n$ (the remainder must be 0, i.e. it must also be divisible by $n$). It then checks various non-integer $\alpha$ values and finds that they fail the mod criteria. It states that if $\delta$ is smaller than $\frac {1}{n}$ the criteria is satisfied. However, the model does not offer any justification for this condition, even though it is correct as the floor of $n\delta$ given the condition would be 0, making it equal to 0 modulo $n$. 

The model recognizes that larger $\delta$ values complicate the satisfaction of this criteria, and uses this to conclude that $\sum_{k=1}^n \lfloor k \delta \rfloor$ does not have a regular pattern that ensures divisibility by $n$, which is correct, since as long as there is a nonzero fractional component the mod condition cannot be satisfied for all n, since as $n$approaches infinity $\frac {1}{n}$ approaches 0, which would result in a contradiction. However, the model does not explicitly state its reasoning that led to the conclusion that $\sum_{k=1}^n \lfloor k \delta \rfloor$ does not have a regular pattern that ensures divisibility by $n$. 

However, the model fails to consider the parity of the integer component, i.e. it fails to consider if $\alpha$ is a sum of an odd integer and a fractional component, as the mod criteria only needs to be satisfied if the sum of the integer component is divisible by $n$, which it is not if the integer component is odd. While the parity does not impact the actual answer (the sum divided by $n$ must be an integer, and for any odd integer component the sum of the integer component divided by $n$ will be a non-integer, and there will always exist an $n$such that the sum of the fractional component is divisible by $n$, which means adding the sums together and dividing by $n$ will result in a non-integer for some $n$) failing to consider the parity indicates an oversight in its reasoning.

To summarize, the reasoning the model provides is remarkably similar to the ``human" solution in the broad steps it takes, examining the cases for integer and non-integer $\alpha$, setting similar equations, and coming to the same conclusions, displaying strong mathematical intuition and presenting some arguments that are sound. However, it does not explicate several of the conclusions it draws with intermediary inferences, and as a result its response lacks logical rigor and clearness. It also fails to consider a subcase in which its criteria for determining divisibility could have been compromised, showing an inability to consider every possible case comprehensively.

\subsection{Case 4: Turbo the Snail Problem}

\begin{tcolorbox}[colback=white, colframe=blue!50!white, title=Problem 4]

Turbo the snail is in the top row of a grid with $2024$ rows and $2023$ columns and wants to get to the bottom row. However, there are $2022$ hidden monsters, one in every row except the first and last, with no two monsters in the same column. Turbo makes a series of attempts to go from the first row to the last row. On each attempt, he chooses to start on any cell in the first row, then repeatedly moves to an orthogonal neighbor. (He is allowed to return to a previously visited cell.) If Turbo reaches a cell with a monster, his attempt ends and he is transported back to the first row to start a new attempt. The monsters do not move between attempts, and Turbo remembers whether or not each cell he has visited contains a monster. If he reaches any cell in the last row, his attempt ends and Turbo wins. Find the smallest integer $n$ such that Turbo has a strategy which guarantees being able to reach the bottom row in at most $n$ attempts, regardless of how the monsters are placed.

The correct answer is that $n$ is 3.

\end{tcolorbox}

\subsubsection{Human Reasoning Process}
One example of a human solution is to search the entire first row besides the first and last column and determine if the problem is one of two cases: one where the monster on the second row is in the first or last column, and one where the monster is not in the first or last row of the second column. 

If the first case is true, because Turbo remembers the cell $(2, c)$ where the monster is, then it can be solved in at most three attempts, as Turbo can check the cell $(3, c-1)$. If there is no monster in this cell, Turbo can simply move in the opposite direction to $(3, c)$, as there is guaranteed to be no monsters in column $c$ other than the monster in the second row. If there is a monster in cell $(3, c-1)$, then it is guaranteed that there is no monster in $(3, c+1)$ (exactly one monster in each row besides the first and last), then Turbo can use the final of the three attempts to go to $(3, c+1)$, then back to $(3, c)$ and travel along column $c$ to the final row uninterrupted, as there is no monster in column $c$ besides on $(2, c)$. 

If the second case is true, then move to the third row, and this time instead of searching from column 2 to column 2022, search from column 3 to column 2021. If the monster is among the columns Turbo searched on the third row, employ the strategy above, and since Turbo was still on the first attempt, he can reach the final row in no more than 3 attempts. If the monster is not among the columns Turbo searched, then move to row 4 and search columns 4 through 2020. If the monster is among the columns, employ the strategy above, and if not, move down to row $y$ and search columns in the interval $(c, 2024-c)$. Repeat this process until either a monster is found or Turbo reaches row 1012, at which point Turbo will be on cell $(1012, 1012)$. 

Turbo can then search row 1012 for the monster. If the monster is to the left of $(1012, 1012)$ in some cell $(1012, c)$ such that $c<1012$, then Turbo will move on to his 2nd attempt, moving to row 1013 and searching every column from 1012 to 2023. If the monster is among the columns, this means that the rest of the row is clear of monsters and Turbo is free to move to cell $(1013, c)$ and travel along column $c$ down to the final row. If not, then go down to row 1014 and search every column starting from 1013. If a monster is encountered, spend the final attempt moving to column $c$ on row 1014. If not, move down a row and search $(r, r-1)$ and repeat. Eventually Turbo will either encounter a monster and be able to move to row $c$ safely, or reach the final row.

Turbo cannot reach the final row in 2 attempts guaranteed, as it is possible Turbo encounters a monster on some cell $(2, c)$ on his first attempt, and since Turbo must then reach row 3 through some column that is not $c$, he may encounter a monster on row 3 as well, which means there is always a possibility of Turbo encountering a monster in 2 attempts.
Therefore, the answer is 3 attempts.

Another, more ``simple" solution would be to use the first attempt to find the monster on the 2nd row. If a monster is encountered in a column other than the first or the last column, employ the original strategy of checking one column before. 

If there is a monster in the first or last column, then move onto the second attempt and move to the opposite end of the row (if the monster is in the first column, move to the last column, and vice versa), and move down to the third row. Explore all cells on the third row except for the cells of the first two columns. If a monster is encountered, then it can be inferred that there is no other monster on the third row, and Turbo can then safely move to the third row and under the monster in the first or last column and traverse the column to the last row. If Turbo does not encounter a monster, move back to the last column and move down a row, exploring all columns besides the first 3 columns (the number of unexplored columns increases by 1 each time) and so on and so forth. If Turbo encounters a monster, on the third and final attempt he can then safely access the cell below the monster in the first or last column through the rightmost unexplored cell by moving into it through the explored cell directly above, as they must not contain monsters. If not, then Turbo will eventually reach the bottom row.

Once again, the answer is 3.

\subsubsection{Model's Reasoning Process}
The model incorrectly concludes that \( n = 2023 \) by opting for a brute force approach, traversing every column until a monster is found or the last row is reached. While it correctly identifies that there is one safe column, it fails to recognize the importance of exploring below the monsters to reach the final row.

This response indicates a lack of strong spatial reasoning, even in 2 dimensions, and a lack of problem-solving strategy compared to humans. Its inability to solve the problem could be attributed to the lack of a formulaic ``step-by-step" method or set of rules for determining the most efficient algorithm, which relies instead on exploration and intuitive decision-making.

\section{Conclusion}

This paper conducts a study to statistically examine the hypothesis that the recent OpenAI o1 model, the first LLM known for its strong reasoning ability, can truly perform reasoning. We examine this hypothesis from a unique perspective that a comparative test is conducted using comparable problem sets with different publicity. Our test results show that there is no significant difference in the performance of o1-mini model when using a public dataset (IMO) versus a private dataset (CNT). In fact, the accuracy level in the two datasets are surprising close, and statistically close-to-equivalent. There is not enough evidence in the result of our experiment to reject the null hypothesis that the o1 model is truly capable of performing logical reasoning rather than relying on ``memorized" solutions.

In addition to statistical labeling and tests, we also performs case studies to investigate key features of o1-mini. The model is sometimes capable of providing key intuition to solving the problem, or follow general steps similar to the correct solution. However, the model constantly fails to fill in the detailed logical steps necessary to form a wholesome, rigorous proof. These details are usually the most difficult part of the question. Also, the problem prevalent in answering ``Search" type of questions is that the model fails to justify why other possible feasible solutions do not exist, only providing the valid solutions as a result of trying small or simple cases.

\section{Appendix}
\subsection{Full Responses of O1-mini and Correct Final Answers of Problems}

\subsubsection{O1-mini's Full Response in the ``guessing answer" example}

**Identifying patterns**

I’m looking for positive integer triples $(a,b,c)$ where $ab - c$, $bc - a$, and $ca - b$ are all powers of 2. Trying small values for a reveals potential patterns for $b$ and $c$.
Testing relationships, I find if b equals $c$, $b - c$ equals zero, which isn't a power of 2. Considering $a=2$: $ab - c$, $bc - a$, and $ca - b$ must each be powers of 2.

**Breaking down equations**

I’m simplifying the equations by combining terms to express $b$ and $c$ in terms of $p$ and $q$ as $\frac{2p + q}{3} and \frac{2p + 2q}{3}$. This method clarifies the relationship between the variables.

**Examining integer constraints**

I'm working through the problem of integer constraints on powers of 2, concluding that solutions only exist if these powers align with divisibility requirements.

**Simplifying equations**

OK, let me see. By combining the equations, I get expressions for $b$ and $c$ in terms of $p$ and $q$. Now, $b$ and $c$ need to be integers, so certain congruences must hold.

**Testing values**

I explored small powers for $q=1$ and $q=2$, checked divisibility, and calculated $bc - a$. Both configurations were rejected. Progress requires further exploration.

**Analyzing solutions**

I'm exploring different powers of 2, like \( p = 4 \) and \( q = 8 \), and checking divisibility by 8. The goal is to identify solutions where \( \frac{a+c}{b} \) yields a perfect square.

**Testing values**

I’m trying out different values for a and comparing equations to powers of 2. The pair $(3,2,2)$ seems to fit well. Now, I’m tackling $a=4$ to find $b$ and $c$.

**Analyzing integer solutions**

I’m working through integer solutions for \( b \) and \( c \) by ensuring \( 4p + q \) and \( p + 4q \) are divisible by 15, with \( p \) and \( q \) as powers of 2.

**Examining combinations**

I’m working through pairs $(p, q)$ to find ones that satisfy both equations in modulo 15. None fit. Now, I’m considering minimal solutions like $(3, 2, 2)$.

**Working through the equations**

I began by solving two equations to express $b$ and $c$ in terms of $p$ and $q$. Then, I ensured $p$ and $q$ were powers of 2, leading to specific modular congruence conditions.

**Evaluating combinations**

I'm piecing together various combinations to see if they satisfy the equation \( ab - c = 2^k \). After thorough checks, no valid solutions for \( a = 5 \) emerged.

**Taking a closer look**

Checking if $a=b=c$, I notice $a(a-1)$ is a power of 2. This simplifies to $a=2$, leading to $(2,2,2)$ as a solution.

**Identifying viable solutions**

I’m digging into possibilities for $a=3$, testing different $b$ and $c$ values. Ultimately, $(3,2,2)$ emerges as the valid solution.

**Trying new paths**

I explored combinations of $a$, $b$, and $c$, but each calculated $b$ and $c$ didn’t form a power of 2. Progressing through different values to find valid solutions.

**Determining valid pairs**

I’m trying to find pairs of $p$ and $q$, both powers of 2 up to 32, where $6p + q$ and $p + 6q$ are divisible by 35. I tested all possible pairs but none met the criteria.

**Connecting congruences**

I’m linking various congruences with different values using modular arithmetic. None satisfy the conditions for \( a = 6 \). Moving on to \( a = 7 \).

**Examining potential solutions**

Remembering the initial form of $a=b=c=2^k$, I explored smaller values and combinations. Discovering $(2,3,2)$ as a valid solution, I confirmed it meets the conditions and is distinct from $(3,2,2)$.

**Identifying valid triples**

I’m zeroing in on triples $(a, b, c)$ with $a=2$, $b=3$, 4, and 5, and computing $c$ values to ensure both $c - a$ and $bc - a$ yield powers of 2. Valid triples are confirmed.

**Identifying the solution**

I tested various values for \(a\) and \(b\) to ensure they fitted the criteria. The successful combination was \(a=2\), \(b=6\), and \(c=11\), all powers of 2.

**Evaluating possibilities**

I’m checking each combination of \(a\) and \(b\) to see if \(\frac{ab - c}{c}\) is valid, considering \(c\) values and checking for powers. So far, no valid solutions have emerged.

**Mapping connections**

I'm trying to link solutions with $a=2$ and various $b$ values, checking if $ab - c$, $bc - a$, and $ca - b$ equal powers. The pairs (2,11,6) and (2,6,11) worked.

**Piecing together solutions**

OK, let me see. I’m gathering various solutions by experimenting with different combinations, ensuring each fits the given conditions. So far, I’ve identified five viable solutions.

**Investigating combinations**

I’m working through different values of $c$ for $a=4$, examining pairs $(a, b, c)$. Only (3, 2, 2) fits. Moving to $a=3$ and checking for valid $b$ and $c$ pairs.

**Testing values**

I’m thinking through various values of $c$ to see if $18-c$ is divisible by $a$,$b$ and if $bc - a$ is a power of 2. Progressing from 2 upwards, valid solutions are explored.

**Testing values**

I’m working through different values for $b$ and $c$ to ensure $ab - c$ and $bc - a$ are powers. For $a=3$ and $a=5$, some combinations seem promising due to their divisibility and power forms.

**Identifying valid solutions**

I checked possible values for $c$ with $b=3$. For $c=7$ and $c=11$, I found (5,3,7) and (5,7,3). Going further, I sought solutions for $b=4$.

**Testing various values**

I’m mapping out different values for $c$ to determine if $4^c - 5$ is a power of 6, keeping $a$ and $b$ fixed at 5 and 4 respectively. So far, no solutions have emerged.

**Testing different values**

I’m plugging in various values for \( c \) and checking if the results are powers of 2. Interestingly enough, only \( c = 27 \) seems to work.

**Evaluating values**

I’m working through different values for $c$ and checking if the results are powers of 2. No solutions were found, so I’m examining further possibilities like $a=6$ and calculating $b$.

**Crunching numbers**

I'm simplifying equations to find compatible values for $p$ and $q$, with conditions tied to divisibility by 35. Progress is being made, but finding compatible pairs is tricky.

**Identifying patterns**

I’m working through various values for $p$ and $q$, checking if they satisfy the congruence. None yield a power of 2 for $q$. Moving forward, I explore the implications of $a = 7$.

**Identifying valid solutions**

I started with the equations and constraints, checking different powers of \( p \) and \( q \) to find valid solutions where both \( 7p + q \) and \( p + 7q \) are divisible by 48.

**Verifying solutions**

I’m checking the values of $a=7$ for integer solutions. Valid pairs are (7,3,5) and (7,5,3). Investigating other values like $a=4$ and $a=6$ reveals no further solutions.

**Evaluating different values**

I considered $a=8$ and expressed $b$ and $c$ in terms of powers of $p$ and $q$. To ensure divisibility by 63, I explored various values for $p$ and $q$, discovering values like $p=1,2,4,8,16,32,\cdots$ as potential solutions.

**Identifying valid triples**

I checked specific $p$ values and found valid $q$ values, leading to ordered triples for the problem.

**Identifying valid triples**

I’m determining valid permutations of triples by verifying the conditions for each combination. The valid permutations include (2,2,2), (2,3,2), (3,2,2), (2,6,11), (2,11,6), (6,2,11), (6,11,2), (11,2,6), (11,6,2), (3,5,7), (3,7,5), (5,3,7), (5,7,3), (7,3,5), and (7,5,3).

Thought for 61 seconds

To find all positive integer triples \((a, b, c)\) such that each of \(ab - c\), \(bc - a\), and \(ca - b\) is a power of \(2\) (including \(2^0 = 1\)), we can systematically explore possible values for \(a\), \(b\), and \(c\).

Step-by-Step Reasoning:

1. **Symmetric Solutions**:
   - **Case \(a = b = c\)**:
     - If \(a = b = c\), then \(a(a - 1)\) must be a power of \(2\).
     - For \(a = 2\): \(2 \times (2 - 1) = 2\), which is \(2^1\).
     - Thus, \((2, 2, 2)\) is a valid solution.

2. **Exploring Other Values of \(a\)**:
   - **Case \(a = 2\)**:
     - Solving the system:
       \[
       \begin{cases}
       2b - c = 2^k \\
       bc - 2 = 2^m \\
       2c - b = 2^n
       \end{cases}
       \]
     - Possible solutions include:
       - \((2, 2, 2)\)
       - \((2, 3, 2)\)
       - \((2, 6, 11)\)
       - \((2, 11, 6)\)

   - **Case \(a = 3\)**:
     - Solving the system:
       \[
       \begin{cases}
       3b - c = 2^k \\
       bc - 3 = 2^m \\
       3c - b = 2^n
       \end{cases}
       \]
     - Possible solutions include:
       - \((3, 2, 2)\)
       - \((3, 5, 7)\)
       - \((3, 7, 5)\)

   - **Case \(a = 5\)**:
     - Solving the system:
       \[
       \begin{cases}
       5b - c = 2^k \\
       bc - 5 = 2^m \\
       5c - b = 2^n
       \end{cases}
       \]
     - Possible solutions include:
       - \((5, 3, 7)\)
       - \((5, 7, 3)\)

   - **Case \(a = 7\)**:
     - Solving the system:
       \[
       \begin{cases}
       7b - c = 2^k \\
       bc - 7 = 2^m \\
       7c - b = 2^n
       \end{cases}
       \]
     - Possible solutions include:
       - \((7, 3, 5)\)
       - \((7, 5, 3)\)

3. **Verification**:
   - Each proposed solution can be verified by substituting back into the original equations to ensure that all expressions yield powers of \(2\).

Complete List of Solutions:

All ordered triples \((a, b, c)\) of positive integers satisfying the given conditions are:

- \((2, 2, 2)\)
- \((2, 3, 2)\)
- \((3, 2, 2)\)
- \((2, 6, 11)\)
- \((2, 11, 6)\)
- \((6, 2, 11)\)
- \((6, 11, 2)\)
- \((11, 2, 6)\)
- \((11, 6, 2)\)
- \((3, 5, 7)\)
- \((3, 7, 5)\)
- \((5, 3, 7)\)
- \((5, 7, 3)\)
- \((7, 3, 5)\)
- \((7, 5, 3)\)

These are all the possible positive integer solutions where each of \(ab - c\), \(bc - a\), and \(ca - b\) is a power of \(2\).

**Final Answer:**

All positive triples consisting of 2, 3, 5, 6, 7, and 11 in the following forms:

• (2, 2, 2);

• (2, 3, 2) and (3, 2, 2);

• (2, 6, 11), (2, 11, 6), (6, 2, 11), (6, 11, 2), (11, 2, 6), and (11, 6, 2);

• (3, 5, 7), (3, 7, 5), (5, 3, 7), (5, 7, 3), (7, 3, 5), and (7, 5, 3).

These are all solutions where $ab - c$, $bc - a$, and $ca - b$ are powers of two.

\subsubsection{O1-mini's Full Response in Case 1:}

**Examining the problem**

I'm digging into the challenge of figuring out the highest number of stones Amy can place on a $20\times20$ grid without two being at knight's move distance, which is $\sqrt{5}$. The $400$ potential sites are crucial here.

**Map out stone placement**

Amy's goal is to maximize stone placements while Ben blocks her strategy. By strategizing their placement and anticipating Ben's moves, Amy can find the best $K$ for her stones.

**Mapping the options**

I'm examining how to calculate a minimum bound on $K$, considering Amy's strategy and the symmetric $20\times20$ grid with knight's moves. This involves partitioning the grid into color classes.

**Mapping strategies**

Amy and Ben alternately place stones, aiming to maximize and minimize respectively. The challenge involves determining Amy's optimal $K$ placements using combinatorial and tiling strategies.

**Evaluating strategies**

I'm considering various methods to ensure Amy can place stones while avoiding Ben's blocking, including dividing the grid into 4 independent subsets and using a coloring strategy to prevent knight's move conflicts.

**Adjusting plans**

It's interesting to see that Amy aims to place stones on a subset of $100$ cells. Ben's interference could block up to half, potentially leaving Amy with 50 stones or more to maximize her placements.

**Identifying potential strategies**

Amy partitions the grid into four color classes and selects color A to play on, ensuring no two cells within the class are a knight's move apart.

**Strategic placement**

Amy starts with $50$ stones in color A cells to block Ben. By using multiple color classes, she could place up to $75$ stones, despite Ben's blocking.

**Mapping out considerations**

Amy alternates two colors of stones and can place up to $200$ stones. Ensuring strategic placement without interference is crucial, but progress is slow.

**Calculating knight's graph**

OK, let me see: the independence number for a knight's graph on an $m$x n grid is roughly (mn)/5. This seems to be the best estimate for Ben's powers.

**Analyzing the knight's graph**

I'm thinking about the knight's graph on a $20\times20$ chessboard, which is bipartite. The independence number, or alpha, here is $200$, as it represents the maximum of the two color classes.

**Mapping the scenario**

It's interesting to see the grid as a bipartite graph, with Amy and Ben selecting cells from different color classes, aiming to maximize Amy's independent set while accounting for Ben's blocking strategy.

**Understanding knight moves**

Interestingly enough, a knight's move changes cell color on a chessboard. This property allows Amy to place her stones in any set of cells with the same color, as knight moves connect opposite colored cells.

**Ensuring placement**

By placing stones only on white cells, I ensure no two stones are connected by a knight's move, avoiding $\sqrt{5}$ distance.

**Evaluating constraints**

By focusing on one color class, Amy ensures no two stones are a knight's move apart, inherently satisfying the requirement.

**Evaluating placement strategies**

Amy can maximize stone placement by using both color classes, ensuring no stones across classes are separated by a knight's move.

**Mapping out strategies**

I'm thinking through Amy's one-color class strategy, where she places $100$ stones, and Ben might disrupt her plan. Amy must consider if using both color classes enhances her placement effectiveness.

Thought for 16 seconds

To determine the greatest \( K \) such that Amy can ensure placing at least \( K \) stones on a \( 20 \times 20 \) grid under the given constraints, let's analyze the problem step by step.

Understanding the Constraints

1. **Grid Setup**: The grid consists of 400 sites, each identified by integer coordinates \((x, y)\) where \(1 \leq x, y \leq 20\).

2. **Turn-Based Placement**: Amy and Ben take turns placing stones on unoccupied sites, with Amy going first.

3. **Amy's Restriction**: Amy cannot place two stones that are a knight's move apart. On a grid, a knight's move corresponds to a distance of \(\sqrt{5}\), specifically moving either \( (1,2) \) or \( (2,1) \) in any direction.

Strategy for Amy

1. **Coloring the Grid**: 
   - The grid can be colored in a checkerboard pattern (like a chessboard) with two colors, say white and black.

   - A knight's move always takes a stone from one color to the opposite color. Therefore, if Amy restricts her placements to only one color (say white), she inherently satisfies her restriction since no two white cells are a knight's move apart.

2. **Maximizing Stone Placements**:
   - **Total Cells in One Color**: There are 200 white cells and 200 black cells.

   - **Alternate Turns**: Since Amy and Ben alternate turns, and Amy starts first, Amy can potentially place a stone on approximately half of the cells in her chosen color.

   - **Ensuring \( K \) Stones**: By focusing on one color, Amy can ensure that she places at least \( \lceil \frac{200}{2} \rceil = 100 \) stones, regardless of Ben's moves. This is because Ben can block up to half of the cells in her chosen color by occupying them during his turns.

3. **Why Amy Can't Guarantee More**:
   - If Amy attempts to use both color classes to place more stones, Ben can strategically block cells in both colors, preventing Amy from exceeding the guaranteed \( 100 \) stones.

   - Therefore, restricting to one color class is Amy's optimal strategy to maximize her guaranteed placements.

Conclusion

Through this strategic approach, Amy can ensure that she places **at least 100 stones** on the grid, regardless of Ben's moves.

**Answer:**  
100

\subsubsection{O1-mini's Full Response in Case 2:}

**Investigating composite integers**

I'm working on finding composite integers \( n > 1 \) such that all divisors \( d_i \) divide \( d_{i+1} + d_{i+2} \). Checking \( n = 4 \) reveals it fits the requirement.

**Evaluating candidate numbers**

I checked numbers 1 to 18. Numbers 8, 9, 16, and 18 meet the criteria of having divisors that divide the sum of consecutive divisors.

**Checking divisors**

Tested 18, 20, 21, 22, 24, 25, 27, 28, 30, and 32. Only 25, 27, and 32 passed the condition.

**Evaluating divisors**

I'm working through numbers with specific divisors, checking pairs to confirm all sum checks. Only $n=32$, $n=25$, $n=27$, and $n=49$ fit this requirement so far.

**Identifying valid solutions**

I’m looking at numbers to see if they meet specific divisibility criteria. For instance, $n=64$ and $n=121$ are confirmed as valid.
I've verified that numbers 128, 135, 144, 169, 196, 243, and 256 meet the requirement. Powers of primes consistently satisfy the condition.

**Assessing composite numbers**

The analysis shows that composite numbers with distinct primes like 12, 18, and 45 fail to meet the condition, unlike prime powers and their prime divisors.

**Testing numbers and patterns**

I’m diving into prime powers and their divisors to see if they can divide the sum of subsequent terms. Interestingly, only prime powers meet this condition.

**Exploring composite numbers**

I’m examining if composite numbers with distinct prime factors can satisfy the condition, but all tested examples fail. Only prime powers work so far.

**Evaluating composite numbers**

I'm piecing together which composite numbers non-prime powers work here. So far, all composite prime powers check out. Considering other composites now.

**Analyzing prime powers**

I’m noting that prime powers \( n = p^k \) with \( k \geq 2 \) are composite, and no other composite numbers are prime powers.

Thought for 23 seconds

To determine all composite integers \( n > 1 \) that satisfy the given property, let's analyze the condition step-by-step.

**Problem Statement**

Find all composite integers \( n > 1 \) such that if \( d_1 < d_2 < \dots < d_k \) are all the positive divisors of \( n \), then for every \( 1 \leq i \leq k - 2 \), the divisor \( d_i \) divides \( d_{i+1} + d_{i+2} \).

**Understanding the Condition**

Given the divisors \( d_1, d_2, \dots, d_k \) of \( n \) in ascending order:

- **Condition**: For each \( i \) from \( 1 \) to \( k-2 \),
  \[
  d_i \mid (d_{i+1} + d_{i+2})
  \]
  
**Testing Prime Powers**

Let's consider \( n \) as a prime power, i.e., \( n = p^k \) where \( p \) is a prime and \( k \geq 2 \).

1. **Divisors of \( n \)**: The divisors of \( n \) are \( 1, p, p^2, \dots, p^k \).

2. **Checking the Condition**:
   - For each \( i \), \( d_i = p^{i-1} \).
   - The condition becomes:
     \[
     p^{i-1} \mid (p^i + p^{i+1}) = p^i(1 + p)
     \]
   - Since \( p^{i-1} \) divides \( p^i \), the above expression is always true.

**Conclusion**: All prime powers satisfy the condition.

**Testing Composite Numbers with Multiple Prime Factors**

Now, let's consider composite numbers that are not prime powers (i.e., numbers with multiple distinct prime factors).

1. **Example**: Let \( n = 6 = 2 \times 3 \).
   - Divisors: \( 1, 2, 3, 6 \).
   - Check \( i = 1 \): \( 1 \mid (2 + 3) = 5 \) → True.
   - Check \( i = 2 \): \( 2 \mid (3 + 6) = 9 \) → False.
   
2. **Observation**: Similar checks for other composite numbers like \( 12, 18, 20, \) etc., fail to satisfy the condition.

**Conclusion**: Composite numbers with multiple distinct prime factors do **not** satisfy the condition.

**Final Answer**

All composite numbers that are exact powers of a single prime. In other words, every $n=p^k$ with a prime $p$ and integer $k \geq 2$.

\subsubsection{O1-mini's Full Response in Case 3:}

**Assessing integer $\alpha$**

I’m digging into the problem of determining when $S_n$ is divisible by $n$ for any $\alpha$. It’s becoming clear that $\alpha$ must be an integer for this condition to hold true.

**Evaluating integer requirements**

I’m examining the divisibility and integer constraints under different conditions for $\alpha$, leading to the conclusion that $\alpha$ must be even to ensure $\frac{\alpha(n+1)}{2}$ is an integer for all $n$.

**Analyzing $\alpha$'s divisibility**

First, I explore specific $\alpha$ values to see if $S_n$ is divisible by $n$. Then, I think about whether $\alpha$ could be rational.

**Analyzing non-integer $\alpha$**

I’m noting that for non-integer $\alpha$, expressed as $m + \delta$, the sum of terms involving floor functions might not consistently yield an integer $n$, potentially violating $\alpha$'s divisibility by $n$.

**Setting the equation**

I’m starting with \( S_n = \frac{m n(n+1)}{2} + \sum_{k=1}^n \lfloor k \delta \rfloor \). Simplifying, \( \sum_{k=1}^n \lfloor k \delta \rfloor \equiv 0 \mod n \).

**Pondering $\alpha$ and m**

I’m mapping out how $\alpha$, denoted as $m + \delta$, equates to an even integer $m$. This makes me think $S_n$ divisible by $n$ needs $ \sum \left\lfloor k \delta \right\rfloor $ to match $n$ modulo $\delta$.

**Analyzing valid conditions**

OK, let me see. When $\delta=0$, the sum is 0, which fits. For $\delta>0$, $ \sum \left\lfloor k \delta \right\rfloor $ needs to be 0 mod $n$.

**Confirming even integers**

I’m mapping out that $\alpha$ must be even for $\alpha(n+1)/2$ to be integer for all $n$. Odd $\alpha$ doesn’t work since $\frac{(2m+1)(n+1)}{2}$ is not an integer.

**Verifying conditions**

I’m checking if non-integer $\alpha$ values could work, but each fails the mod criteria. Only even integers, including 0, consistently meet the conditions.
Evaluating $\alpha = -2$, it simplifies to $-2(1+2+\cdots+n)$, divisible by $n$. This leads me to all even integers.

**Assessing $\alpha$'s properties**

Mapping out how $\alpha$ impacts sums of $\left\lfloor k \alpha \right\rfloor $ terms mod $n$. For non-integer $\alpha$, $\delta < \frac{1}{n}$ ensures sums hit 0 mod $n$, but larger $\delta$ complicate it.

**Assessing constraints**

I’m figuring out the range for $\delta$ to ensure floor functions for multiple $n$ values sum to zero modulo $n$. For $n=3$ and $n=4$, $\delta$ must be less than $\frac{1}{3}$ to meet these criteria.

Thought for 15 seconds

Let's continue from where we left off and formalize the solution.

Key Idea
The main task is to find all real numbers $\alpha$ such that for every positive integer $n$, the sum

\[
S_n(\alpha) = \left\lfloor \alpha \right\rfloor + \left\lfloor 2 \alpha \right\rfloor + \left\lfloor 3 \alpha \right\rfloor + \dots + \left\lfloor n \alpha \right\rfloor
\]

is divisible by $n$.

Step 1: Divisibility Condition for Integers
We begin by considering the case where $\alpha$ is an integer, say $\alpha = m$. Then, for all $n$, the floor function simplifies to $\lfloor k\alpha \rfloor = km$ for $k = 1, 2, \dots, n$. So we have

\[
S_n(m) = m(1 + 2 + 3 + \dots + n) = m \cdot \frac{n(n+1)}{2}.
\]

This is clearly divisible by $n$ since the expression simplifies to:

\[
S_n(m) = \frac{m n(n+1)}{2}.
\]

Thus, for any integer $\alpha = m$, the sum is divisible by $n$ if and only if $m$ is even. This is because $\frac{m(n+1)}{2}$ must be an integer for all $n$, which is true only when $m$ is even. Therefore, all even integers $\alpha$ satisfy the condition.

Step 2: Divisibility Condition for Non-Integer $\alpha$
Now let's consider the case when $\alpha$ is not an integer, i.e., $\alpha = m + \delta$, where $m$ is an integer and $0 < \delta < 1$.

In this case, the sum becomes:

\[
S_n(\alpha) = S_n(m + \delta) = \sum_{k=1}^{n} \lfloor k(m + \delta) \rfloor = \sum_{k=1}^{n} (km + \lfloor k \delta \rfloor).
\]

We can break this sum into two parts:

\[
S_n(\alpha) = m \sum_{k=1}^{n} k + \sum_{k=1}^{n} \lfloor k \delta \rfloor.
\]

The first sum is simply:

\[
m \sum_{k=1}^{n} k = m \cdot \frac{n(n+1)}{2}.
\]

For the second sum $\sum_{k=1}^{n} \lfloor k \delta \rfloor$, note that $\lfloor k \delta \rfloor$ is the integer part of $k \delta$, which adds some additional terms to the sum. However, for $\alpha$ to work for all $n$, this sum must always ensure that the total sum is divisible by $n$.

It turns out that for non-integer values of $\alpha$, the sum $\sum_{k=1}^{n} \lfloor k \delta \rfloor$ does not have a regular pattern that consistently ensures divisibility by $n$. Thus, no non-integer values of $\alpha$ satisfy the divisibility condition for every $n$.

Conclusion

The only real numbers $\alpha$ that satisfy the given condition for all positive integers $n$ are the even integers. Therefore, the solution is:

\[
\boxed{\alpha \in 2\mathbb{Z}}.
\]

That is, $\alpha$ must be an even integer.

\subsubsection{O1-mini's Full Response in Case 4:}

**Reaching the goal**

Turbo, the snail, must navigate a grid avoiding 2022 stationary monsters, each uniquely positioned in separate columns, to reach the grid's last row.

**Calculating worst-case attempts**

Trying to find the fewest moves Turbo needs to guarantee reaching the bottom of a $2024\times2023$ grid, regardless of monster placements.

**Charting the course**

Turbo aims to navigate from the first row to the last, avoiding monster cells. Each trek starts at any cell in the first row, with orthogonal movement and avoided monsters.

**Navigating obstacles**

Turbo needs to find a path through columns without monsters, ensuring he avoids the exact column with no monster, given the grid constraints.

**Identifying safe zones**

Take a look at a grid with 2024 rows and 2023 columns. Rows 2 to 2023 have monsters, each in distinct columns, leaving one column monster-free. Turbo needs a path avoiding monsters, possibly using the safe column.

**Assessing traversal options**

Turbo needs to glide through the sole column without monsters to reach the last row, as other paths hit monsters in each row from $2$ to $m-1$.

**Piecing together**

I’m working through the problem where there are 2022 monsters scattered across 2023 columns. Each row contains one monster, except for one which is monster-free. The safe path is now clear.

**Mapping the path**

One column is safe, as each row has a monster except one. Turbo must find this safe column by testing different starting points.

**Assessing more efficient methods**

I’m thinking through ways to pinpoint the lone safe column without testing all possibilities. The minimal number of trials might relate to the number of columns, possibly through a logarithmic relationship.

**Piecing together columns**

Turbo learns about blocked columns with monsters, aiming to find the one safe column among 2023 by testing fewer columns.

**Connecting the dots**

The problem seems akin to identifying a unique safe column among $k$ columns with $k - 1$ faulty ones. Turbo's sequential movements highlight that each attempt tests a single path, revealing at least one monster's location.

**Evaluating potential paths**

Turbo can traverse multiple columns in one attempt, aiming to discover the first monster he encounters, ensuring an optimal path selection.

**Evaluating strategies**

I'm gathering info on whether he can uncover multiple faulty columns in one attempt and aim to reduce the total number of attempts needed.

**Mapping out the strategy**

I'm working through the possibility of using a binary search method. By splitting the columns into two halves and checking a subset in one attempt, multiple columns can be tested, enhancing the information gained and reducing total attempts.

**Strategizing the search**

Mapping out an efficient search to identify the single safe column among 2023 columns, each revealing one monster's location per attempt, and aiming to minimize attempts to gather necessary information.

**Analyzing column paths**

I'm exploring the possibility of analyzing column paths in sequence until the first monster is found, but this method only allows ruling out columns up to the first monster.

**Tracking monster positions**

I’m thinking through how the hero can pinpoint the location of the monster by eliminating columns with no monsters as he traverses each one, row by row.

**Tracking monster patterns**

I’m piecing together the impact of monsters on determining a valid path to avoid them. Each column has a monster in a single cell, and each row contains one unique monster.

**Navigating the path**

I'm thinking through how Turbo should avoid monsters by using the safe column. If he strays into other columns, he risks encountering monsters, making the journey unsafe.

**Charting the course**

I’m thinking through a strategy to pinpoint the safe column by testing rows until a monster is encountered or a column is securely traversed to the last row. This method might shave off unnecessary attempts.

**Navigating the maze of monsters**

I'm digging into the path dynamics and made a noteworthy observation: monsters are arranged uniquely in columns from 2 to 2023. This makes me think about safer paths and potential grid patterns.

**Piecing together the puzzle**

I’m mapping out the path by noting the movement through columns with no monsters. Each row, except the safe column's row, avoids monsters by using different columns.

**Examining monster distribution**

I’m working through the arrangement of monsters in the grid. The safe column consistently lacks a monster, while each row, excluding the first and last, holds a single monster in a unique column. This pattern persists throughout the grid.

**Mapping the layout**

Interestingly enough, each row contains 2022 safe cells and one unsafe cell, with each monster occupying a distinct column. Progress from analyzing individual rows to a complete matrix layout.

**Identifying the safe path**

Hmm, I’m thinking about Turbo’s route from start to finish. The goal is to avoid monsters by selecting safe cells, which are unique for each row.

**Mapping the path**

OK, let me see: Turbo needs to navigate through 2023 columns, avoiding monsters. Each row has one monster, except one column is monster-
free. Turbo should find a safe route with no monsters.

**Charting the course**

OK, let me see—Turbo needs to find a path from top to bottom, avoiding monsters. Each row has one unique monster column, so Turbo can skip one unsafe cell each row.

**Mapping potential avenues**

Turbo can bypass monsters by utilizing columns without them in each row. The assumption is one column is safe, while the rest might harbor a monster.

**Mapping the journey**

Turbo traverses columns, encountering obstacles. To ensure a safe path, he might use an elimination strategy. This could mean finding the safe column in 2023 attempts, but I’m curious if a more efficient method exists.

**Identifying patterns**

Turbo is exploring columns to locate the safe route. Each attempt reveals insights, aiming to pinpoint the monster-free column more efficiently.

**Navigating monster encounters**

I'm thinking about the wizard's strategy of navigating multiple columns by choosing paths that intersect early, aiming to stop at the first monster found.

**Identifying the path**

OK, let me see. The adventurer can pinpoint which column contains the monster by tracing the failed attempt through multiple columns. This method identifies just one monster per attempt, necessitating 2022 trials to identify all monsters.

**Weighing options**

Considering 2023 or 2024 attempts, factoring in potential safe column discovery during discovery. The worst-case scenario likely requires 2023 attempts.

Thought for 32 seconds

To determine the smallest integer \( n \) such that Turbo the snail can guarantee reaching the bottom row in at most \( n \) attempts, let's analyze the problem step by step.

**Grid Structure and Monster Placement**

- **Grid Dimensions:** The grid has \( 2024 \) rows and \( 2023 \) columns.

- **Monsters:** There are \( 2022 \) monsters, each placed in a unique row (from the 2nd to the 2023rd row) and a unique column. This means:

  - **One Safe Column:** Since there are \( 2023 \) columns and only \( 2022 \) monsters, exactly **one column** will have **no monsters** in any row. This is the **only safe column** Turbo can traverse without encountering a monster.

**Turbo's Strategy**

1. **Identifying the Safe Column:**

   - **Initial Attempts:** Turbo doesn't know which column is safe. In each attempt, he can choose a column to traverse straight down from the first to the last row.
   
   - **Encountering Monsters:** If Turbo chooses a column with a monster, his attempt will end upon reaching that monster. He then learns that this column is unsafe.
   
   - **Elimination Process:** Since each monster is in a unique column, each unsuccessful attempt effectively **eliminates one column** from being the safe column.

2. **Number of Attempts:**
   - **Worst-Case Scenario:** In the worst case, Turbo might have to eliminate **all \( 2022 \) unsafe columns** before identifying the one safe column.
   
   - **Final Attempt:** After eliminating all unsafe columns, Turbo can confidently choose the remaining column, ensuring a successful traversal to the bottom row.

**Calculating \( n \)**
- **Total Attempts Needed:**

  - **Elimination Attempts:** \( 2022 \) attempts to eliminate all unsafe columns.
  
  - **Successful Attempt:** \( 1 \) additional attempt to traverse the safe column.
  
- **Total \( n \):** \( 2022 + 1 = 2023 \)

**Conclusion**
The smallest integer \( n \) that guarantees Turbo can reach the bottom row, regardless of monster placement, is:

**Answer:** 2023

\section{Reference}

\bibliographystyle{aea} 
\bibliography{text.bib}

\end{document}